\documentclass[10pt,twocolumn,letterpaper]{article}

\usepackage{cvpr}              
\usepackage[ruled,linesnumbered]{algorithm2e}
\usepackage{multirow}
\usepackage{booktabs}
\usepackage{adjustbox}
\usepackage{colortbl}
\usepackage{color}
\usepackage{xr}
\usepackage[dvipsnames]{xcolor}

\definecolor{cvprblue}{rgb}{0.21,0.49,0.74}
\usepackage[pagebackref,breaklinks,colorlinks,citecolor=cvprblue]{hyperref}

\def\paperID{9210} 
\def\confName{CVPR}
\def\confYear{2024}

\title{Robust Transductive Few-shot Learning via Joint Message Passing and Prototype-based Soft-label Propagation}

\author{Jiahui Wang, Qin Xu\footnotemark[1], Bo Jiang\footnotemark[1], and Bin Luo\\
Computer Science and Technology, Anhui University, China\\
{\tt\small wjh@stu.ahu.edu.cn,\{xuqin,jiangbo,luobin\}@ahu.edu.cn}
}

\begin{document}
\maketitle
\footnotetext[1]{Corresponding authors.} 
\begin{abstract}Few-shot learning (FSL) aims to develop a learning model with the ability to generalize to new classes using a few support samples. 
For transductive FSL tasks, prototype learning and label propagation methods are commonly employed.  
Prototype methods generally first learn the representative prototypes from the support set and then determine the labels of  queries based on the metric between query samples and prototypes. 
Label propagation methods try to propagate the labels of support samples on the constructed graph encoding the relationships between both support and query samples. 
This paper aims to integrate these two principles together and develop an efficient and robust transductive FSL approach, termed Prototype-based Soft-label Propagation (PSLP). 
Specifically, we first estimate the soft-label presentation for each query sample by leveraging prototypes.
Then, we conduct soft-label propagation on our learned query-support graph. 
Both steps are conducted progressively to boost their respective performance. 
Moreover, to learn effective prototypes for soft-label estimation as well as the desirable query-support graph for soft-label propagation, we design a new joint message passing scheme to learn sample presentation and relational graph jointly. 
Our PSLP method is parameter-free and can be implemented very efficiently.
On four popular datasets, our method achieves competitive results on both balanced and imbalanced settings compared to the state-of-the-art methods. \textcolor{blue}{The code will be released upon acceptance.}

\end{abstract}    
\section{Introduction}
\label{sec:intro}
Few-shot learning (FSL), which aims to develop a learning model with the generalization ability  to new classes by using a few support
samples, has attracted considerable attention \cite{survey1,pn,rn} in recent years. 
FSL tasks are usually set in both inductive and transductive manners. 
In inductive FSL, each query example is assumed to be classified independently of other examples at inference time.
In contrast, transductive FSL methods infer the label jointly for all samples in the query set and usually demonstrate better performance than inductive FSL.
In this paper, we focus on the transductive FSL task. 

In the past ten years, many methods have been developed for transductive FSL tasks \cite{simple,laplacian,ilpc,putm,protolp,tim,am,epnet,tpn}. 
Overall, these methods can be roughly categorized into two types, i.e., prototype-based and label propagation-based methods. 
Prototype-based methods \cite{tim,tafssl,as,bdcspn,putm,ease} generally first learn a representative prototype for each category and then determine the label of each query sample by using the distance metrics between it and prototypes. 
For example, 
Boudiaf \etal \cite{tim} propose to use class prototypes as the initial parameters of the classifier and optimize them by maximizing mutual information. Simon \etal \cite{as} suggest to project features into a subspace to obtain better prototype representations. Tian \etal \cite{putm} propose to obtain a refined prototype feature by obtaining a probability distribution matrix via conditional transport. 
The main benefit of prototype-based methods is that 
they generally perform robustly w.r.t class-imbalanced transductive FSL tasks. 
However, one main limitation of this type of methods is that they usually fail to fully exploit the relationships of query samples for their label prediction. 

In contrast, label propagation-based transductive FSL methods try to propagate the labels of support samples on the constructed graph encoding the relationships between support and query samples, which thus can fully exploit the relationships of both support and query samples.  
%
For instance, Rodríguez \etal \cite{epnet} propose to simultaneously use label propagation for feature embedding and classification. Lazarou \etal \cite{ilpc} introduce iteratively graph updating and label propagation. Fu \etal \cite{am} suggest incorporating learnable parameters into the graph structure and optimizing it with a loss function after propagation. 
However, although label propagation-based methods can obtain better performance, they usually perform insensitively w.r.t the class-imbalanced transductive FSL problem. 

Our aim in this paper is to integrate the above
two principles together and develop an efficient and robust transductive FSL approach termed Prototype-based Soft-label
Propagation (PSLP). 
To be specific, we first estimate a soft-label presentation for each query sample by leveraging prototype representation. Then, we conduct soft-label propagation
on the query-support graph (QSGraph). 
Both prototype representation and 
soft-label learning are conducted progressively to boost their respective performance. 
Moreover, to learn effective prototype representation for soft-label estimation as well as QSGraph for soft-label propagation, we design a new joint
message passing (JMP) scheme to jointly learn sample presentation and QSGraph structure. 
Overall, the main advantages of the proposed approach are three aspects. 
(1) \textbf{Effectiveness}: The proposed  PSLP method simultaneously integrates the merits of label propagation and prototype-based metric principle with our JMP module to refine prototype representation and graph construction. Thus, it addresses transductive FSL tasks effectively. 
(2) \textbf{Robustness}: Our method conducts soft-label propagation across all samples, allowing for the adaptive adjustment of the propagation strength for each class to fit the class-imbalanced setting, which thus performs robustly w.r.t class-imbalanced problems. 
(3) \textbf{Efficiency}: The proposed method does not involve any learnable parameters, which can be implemented very efficiently when compared with many other related works, as validated in Experiments. 

Note that recent work ProtoLP \cite{protolp} is most related to our work. It also combines prototype learning and label propagation. 
However, our PSLP is significantly different from ProtoLP as follows. 
First, ProtoLP mainly exploits prototypes for graph construction and feature representation and needs a linear projection to obtain the final label prediction. 
In contrast, our PSLP leverages prototypes for soft-label generation explicitly and thus is much lighter than ProtoLP. 
Second, ProtoLP propagates the `known' labels of support samples while our PSLP aims to propagate soft labels for all samples. 
Third, we further design a new JMP module to jointly learn sample presentation, prototype representation, and QSGraph structure for our soft-label propagation process. 
We will show in experiments that our PSLP performs obviously better than ProtoLP on both learning accuracy and efficiency. 


Overall, the main contributions of this paper are summarized as follows, 
\begin{itemize}
    \item We propose a new general transductive FSL scheme by combining prototype representation with label propagation and developing a novel soft-label propagation approach.
    
    \item  We design a novel joint message propagation, which fully leverages the global information between all samples, resulting in optimized sample feature mappings and robust graph structures.
   
    \item We demonstrate the effectiveness, robustness and efficiency of our method on four commonly used benchmarks, achieving competitive results under different settings compared to SOTA.
\end{itemize}

\section{Related Works}
Few-shot learning aims to train a model that can be generalized to new classes by a small number of labeled samples. Existing works often employ meta-learning strategies \cite{metalearn1,metalean2} to train networks. Metric-based \cite{pn,rn,matching}, optimization-based \cite{optim1,optim2,optim3}, and data-based \cite{easy,freelunch,vari} methods are the main research directions of current meta-learning methods. The vast majority of these methods focus on the inductive setting, where the query example is classified independently of the others.
Our method focuses on transductive few-shot learning \cite{tpn,epnet,simple,laplacian}, which refers to simultaneously using samples with both labeled and unlabeled labels for inference on the basis of few-shot learning, requiring the network to effectively utilize the intrinsic connections between all samples. Existing transductive few-shot learning methods can be roughly classified into Prototype-based Methods and Label propagation-based methods, respectively. Below, we introduce these two types of methods.

\noindent\textbf{Prototype-based Methods.}
In transductive few-shot learning, prototype-based methods typically estimate class centers as prototypes and measure the distance between samples and prototypes for classification \cite{pn}. To enhance the feature representation of prototypes, Simon \etal \cite{as} suggest mapping features to a subspace. Lichtenstein \etal \cite{tafssl} reduce the distribution difference between novel and base classes by applying ICA and PCA for dimensional reduction to samples, leading to more accurate prototype estimation. Liu \etal \cite{bdcspn} correct prototypes using pseudo-labels assigned to samples. Hu \etal \cite{ptmap} and Zhu \etal \cite{ease} propose a classifier to update the prototype iteratively by sinkhorn distance \cite{ot} with the constrain of class-balanced prior.
Some methods optimize parameters during testing to obtain improved prototype feature representations. For instance, Boudiaf \etal \cite{tim} initialize prototype features as the initial parameters of the classifier and optimize them by maximizing mutual information and minimizing cross-entropy loss. Tian \etal \cite{putm} utilize a learnable conditional transport metric to optimize label assignment matrices iteratively, obtaining more accurate prototype representations.
Prototype-based methods generally perform robustly on class-imbalanced transductive FSL tasks but are weak in fully exploiting the global relationships of all samples for label prediction.

\noindent\textbf{Label Propagation-based Methods.}
Label propagation \cite{lp,lp2} is widely used in transductive few-shot learning \cite{tpn,epnet,ilpc,am} that leverages the correlations between sample features to construct graph structures and propagates labels from labeled samples to unlabeled samples. For instance,
Rodríguez \etal \cite{epnet} utilize label propagation for both feature mapping and label prediction. 
Lazarou \etal \cite{ilpc} assume a uniform distribution of query samples across classes and optimize label propagation results with optimal transport \cite{ot}. 
Fu \etal \cite{am} propose to optimize learnable class centers and biases through backpropagation during label propagation to mitigate the impact of imbalanced setting. However, few works integrate the design of prototype-based methods and label propagation-based methods. Zhu \etal \cite{protolp} propose to build the graph by prototypes and optimizing the projection of labels by label propagation.
However, these works typically only use support samples as the label propagation source, 
it defaults to a class-balanced prior assumption, making it difficult to handle cases where the number of query samples per class is imbalanced \cite{atim}.
Our method propagates soft labels by all samples, which can adapt well to imbalanced setting, and the simplified prototype-based propagation steps also result in a more efficient inference speed.
\begin{figure*}[t]
    \centering
    \includegraphics[width=\linewidth]{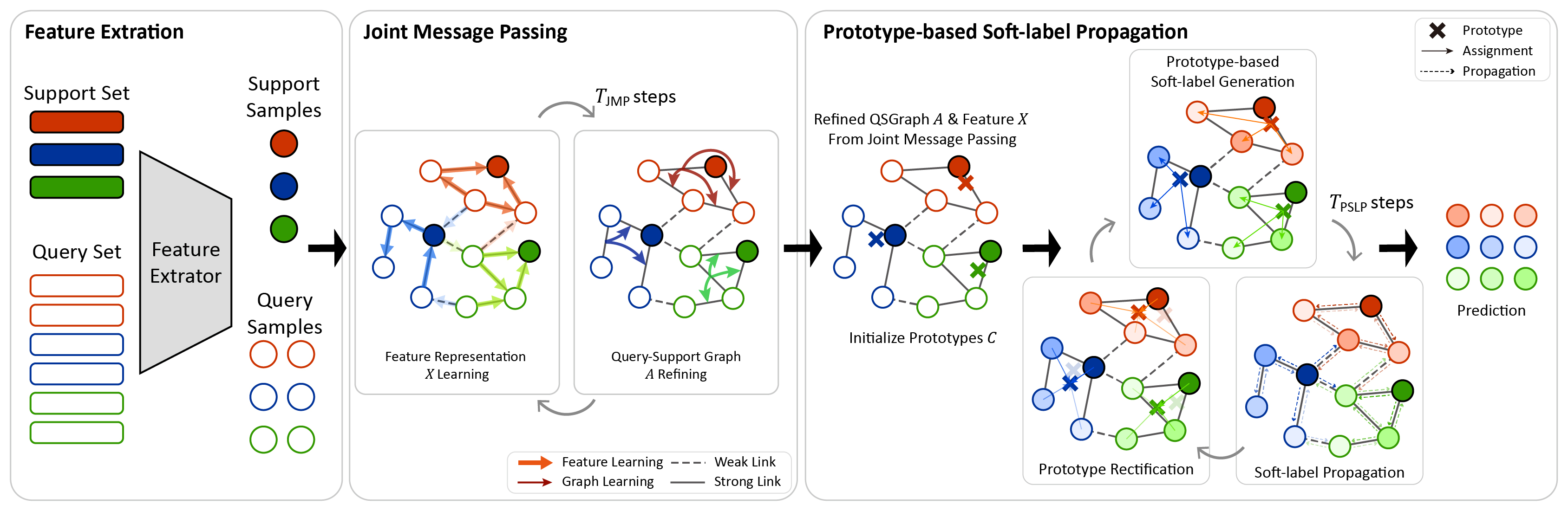}
    \caption{The process of the proposed method. First, we optimize the feature representation and the query-support graph by joint message passing. Then, we use the refined features and graph for prototype-based soft-label propagation to alternately update the labels and prototypes, and output the prediction results.}
    \label{fig:network}
\end{figure*}

\section{Methodology}

\textbf{Problem Formulation.} Under the few-shot scenario, let us assume that we have a large labeled base dataset $\mathcal{D}_{base} = {\{(\boldsymbol{x}^b_i,y^b_i)\}}_{i=1}^{n_{base}}$ with $N_{base}$ classes, where $n_{base}$ denotes the number of base class samples, and $\boldsymbol{x}_i^b$ and $y_i^b$ represent the sample features extracted by the backbone network $f_\theta$ and their corresponding labels, respectively. During the training phase, we train $f_\theta$ on $\mathcal{D}_{base}$ with the supervision of labels. Another dataset of novel classes $\mathcal{D}_{novel} = {\{(\boldsymbol{x}_i,y_i)\}}_{i=1}^{n_{novel}}$ are provided for evaluating classification performance. Here, the novel class labels $y_i$ are categories that the network has never encountered before and satisfy $N_{novel}\cap N_{base} = \varnothing$. 
During the inference phase, for an $N$-way $K$-shot task, we randomly sample support set $\mathcal{S} = {\{(\boldsymbol{x}^s_i,y_i)\}}_{i=1}^{NK} $ and query set $\mathcal{Q} = {\{\boldsymbol{x}^q_i\}}_{i=1}^{M} $ from $\mathcal{D}_{novel}$ to construct multiple tasks, indicating the support set contains $N$ unseen categories, each with $K$ labeled samples, and the $M$ unlabeled samples in the query set all belong to these $N$ categories. We focus on the transductive setting, where we employ $f_\theta$ to extract features and simultaneously utilize all samples from $\mathcal{S}$ and $\mathcal{Q}$, and predict labels $\tilde{y}$ of the samples in $\mathcal{Q}$.

\textbf{Overview of Method.}
In this section, we propose our Prototype-based Soft-label Propagation (PSLP) method for the transductive FSL problem. The proposed method mainly contains 1) Joint Message Passing to learn effective prototype representation and construct the query-support graph (QSGraph) and 2) Soft-label Propagation on QSGraph to achieve label prediction for query samples. 
The diagram of our proposed approach is shown in Fig.~\ref{fig:network}. 

\subsection{Joint Message Passing}
In order to leverage structural information among all samples in both support and query set, we propose Joint Message Passing (JMP) to refine sample representation and graph structure simultaneously. 
Specifically, we first concatenate support and query samples to obtain the initial sample features $X^{(0)}$ as:
\begin{equation}
    X^{(0)} = \mathrm{Concat}([\boldsymbol{x}^s_1,\cdots,\boldsymbol{x}^s_{NK}],[\boldsymbol{x}^q_1,\cdots,\boldsymbol{x}^q_M])
\end{equation}
Using $X^{(0)}$, we then construct an initial query-sample graph (QSGraph) as:
\begin{align}
    A^{(0)}_{ij} &= \mathrm{exp}\big(-\gamma\|\boldsymbol{x}^{(0)}_i-\boldsymbol{x}^{(0)}_j\|_2^2\big)
\end{align}
where $\gamma$ represents the scaling factor. 
Then, we jointly update the sample feature representation $X$ and QSGraph $A$ by progressively repeating the following two steps.

\noindent
\textbf{Step1: Feature Representation Learning.}
Inspired by work \cite{agc}, we employ a message passing mechanism to learn each sample feature representation by aggregating the information of its neighboring samples on QSGraph. Specifically, the optimized features can be achieved by using the message passing as follows:
\begin{equation}
\label{eq:featurerefine}
    X^{(t)} \leftarrow (I+L^{(t-1)})^kX^{(t-1)}
\end{equation}
where $t=1, 2 \cdots T_{\mathrm{JMP}}$ is the iteration step, $L^{(t-1)}=D^{-\frac{1}{2}}A^{(t-1)}D^{-\frac{1}{2}}$ and $D = \mathrm{diag}(\sum_{j}^{NK+M}a_{ij})$ represents the degree matrix. $k\geq 1$ is a hyperparameter to control the size of the receptive field. 
Using Eq.\ref{eq:featurerefine}, each sample learns its context-aware representation by aggregating the information from its $k$-hop neighbors. 

\noindent
\textbf{Step2: QSGraph Refining.}
Based on the above learned representation $X^{(t)}$, we can refine the QSGraph structure $A^{(t)}$ to capture the relationships between all query and sample samples. 
To avoid the over-smoothing issue in the following label propagation  process, we adopt nearest neighbor graph building strategy, i.e., each sample node only connects to its $B$-nearest neighbors as follows: 
\begin{equation}
\label{eq:graphrefine}
    A^{(t)}_{ij}\leftarrow\begin{cases}\mathrm{exp}(-\gamma\|\boldsymbol{x}^{(t)}_i-\boldsymbol{x}^{(t)}_j\|^2_2) & \mathrm{if}\ \boldsymbol{x}^{(t)}_j\in \mathrm{NN}_B(\boldsymbol{x}^{(t)}_i)\\ 0& \mathrm{otherwise}\end{cases}
\end{equation}
where $\mathrm{NN}_B(\boldsymbol{x})$ denotes the set of the $B$-nearest neighbors of sample $\boldsymbol{x}$. 

In practice, the above step 1 and step 2 are alternatively implemented to boost their respective performance. 
After $T_{\mathrm{JMP}}$ iteration, the refined sample feature representation $X$ and QSGraph structure $A$ are used for the following prototype representation and soft-label propagation. We set $T_{\mathrm{JMP}}$ to 1 for efficiency consideration.
Note that the whole process of the above JMP does not involve any learnable parameters, ensuring efficient inference in practical applications. 
 
\subsection{Soft-label Propagation}
To integrate prototype-based method with label propagation, we propose  Prototype-based Soft-label Propagation (PSLP). 
Overall, the proposed PSLP module involves 1) generating soft-label representations for query samples, 2) performing soft-labels on QSGraph to refine/predict the final labels for queries. 
These two processes are progressively performed,   
leading to stable propagation results. The soft-label propagation treats all samples as the propagation source simultaneously and adaptively controls the strength of label propagation for each class, which thus can adapt well to class-imbalanced setting, as validated in Experiments. 
To be specific, we first define the initial prototypes by using the sample features obtained from the above JMP module as follows:
\begin{equation}
\label{eq:initproto}
    \boldsymbol{c}^{(0)}_n = \frac{1}{K}\sum\nolimits_{\boldsymbol{x}'_s\in \mathcal{S}_n}\boldsymbol{x}'_s
\end{equation}
where $X'=\{\boldsymbol{x}'_1, \boldsymbol{x}'_2\cdots \boldsymbol{x}'_{NK}\}$ denotes the learned sample representations of the above JMP module and $\mathcal{S}_n$ denotes the samples of class $n$ in the support set. 
Then, we alternately update prototypes $C^{(0)}=\{\boldsymbol{c}^{(0)}_1,\boldsymbol{c}^{(0)}_2\cdots,\boldsymbol{c}^{(0)}_N\}$ and the label matrix $Z$ by using soft-label propagation.

\noindent
\textbf{Step1: Prototype-based Soft-label Generation.}
To involve all samples in label propagation, we first generate soft-labels to the query samples based on the prototypes. The soft-labels of query samples $\tilde{Z}^{(t)}_{q}\in \mathbb{R}^{M\times N}$ can be obtained based on their similarities to prototypes $C^t$ as: 
\begin{equation}
\label{eq:query}
    \tilde{Z}_{qn}^{(t)} = \frac{\mathrm{exp}(-\gamma\|\boldsymbol{x}_q-
\boldsymbol{c}^{(t)}_n\|^2_2)}{\sum_n^N \mathrm{exp}(-\gamma\|\boldsymbol{x}_q-
\boldsymbol{c}^{(t)}_n\|^2_2)}\quad 
\end{equation}
Since $\sum_n \tilde{Z}_{qn}^{(t)}=1$ and $\tilde{Z}_{qn}^{(t)}\geq 0$, we call them as soft-label representation.  
For support samples, since their labels are known, we simply obtain $Z'^{(t)}_{s} \in \mathbb{R}^{NK\times N} $ of them 
from the known labels $y_s$ as follows:
\begin{equation}
\label{eq:support}
    Z'^{(t)}_{sn} = \mathrm{OneHot}(y_s)= \begin{cases}
 1 & \mathrm{ if }\ y_s= n\\
 0 & \mathrm{otherwise}\end{cases}\quad 
\end{equation}
The label matrix  $Z^{(t)}=[Z^{'(t)},\tilde{Z}^{(t)}]\in \mathbb{R}^{(NK+M)\times N}$  is obtained by concatenating the label matrices of both support and query samples.

\noindent
\textbf{Step2: Soft-label Propagation.}
After obtaining the label matrix $Z^{(t)}$generated under the guidance of prototypes, we then 
conduct label propagation on the above constructed QSGraph. 
Inspired by work \cite{lp}, the propagation result is obtained via a closed-form solution as:
\begin{equation}
\label{eq:propagation}
    Z^{(t+1)}\leftarrow (I-\alpha L)^{-1}Z^{(t)}
\end{equation}
where
$L=D^{-\frac{1}{2}}AD^{-\frac{1}{2}}$ denotes the laplacian matrix of the refined QSGraph and parameter $\alpha\in [0,1)$ represents the propagation strength. In contrast to vanilla label propagation relying solely on the labels of support samples, our method utilizes all samples as propagation sources to obtain more robust propagation results. Finally, we normalize the obtained label matrix as
\begin{equation}
\label{eq:normalize}
    Z_{qn}^{(t+1)} =\frac{Z_{qn}^{(t+1)}}{\sum_n^NZ_{qn}^{(t+1)}}
\end{equation}

\noindent
\textbf{Step3: Prototype Rectification.}
After we obtain the more accurate label prediction via the above propagation process, we can update prototypes via the learned soft-labels. 
Specifically, we update the prototype proportionally to ensure the stability as:
\begin{equation}
\label{eq:update}
    C^{(t+1)} = (1-\beta) C^{(t)} + \beta (Z^{{(t+1)}})^\top X
\end{equation}
where $C^{(t)}=\{\boldsymbol{c}^{(t)}_1,\boldsymbol{c}^{(t)}_2\cdots,\boldsymbol{c}^{(t)}_N\}$ and hyperparameter $\beta$ balances two terms. 

By alternately conducting the above steps for $T_\mathrm{PSLP}$ times, both prototypes and label matrices can be accurately estimated. We finally use the output label matrix $Z^{T_{\mathrm{PSLP}}}_q$ to predict the labels $\tilde{y}_q$ of the query samples as: 
\begin{equation}
    \tilde{y}_q = \mathrm{arg}\max_n(Z_{qn}^{T_{\mathrm{PSLP}}})
\end{equation}
The overall process is outlined in Algorithm 1. 
It is worth noting that the QSGraph is obtained through JMP and remains unchanged in PSLP, so the propagation matrix $P=(I-\alpha L)^{-1}$ in Eq.\ref{eq:propagation} can be obtained outside the iterations to improve inference speed. 
Based on the above progressive learning, both prototypes and label propagation  can be iteratively optimized to boost their performance. Also, for class-imbalanced setting, according to Eq.\ref{eq:propagation}, each query sample propagates its soft-label, allowing the propagation strength of each category to adaptively adjust based on the number of query samples. Moreover, our PSLP method is parameter-free (only contains some hyperparameters) and has simple update steps, typically converging in only 5 to 10 iterations and resulting in high efficiency. 

\begin{algorithm}

\caption{PSLP algorithm}
\label{alg1}
\SetCommentSty{text}
\KwIn{Features $X=\{\boldsymbol{x}_{1}^s,..,\boldsymbol{x}_{NK}^s,..{\boldsymbol{x}_{NK+M}^q}\}$, Support labels $Y=\{y_1,\cdots,y_{NK}\}$, Hyperparamters $k,B,\alpha,\beta$}
\KwOut{Query Labels $\tilde{y}_q$}
\For{$t=1:T_{\mathrm{JMP}}$}{
\tcp{\textcolor{violet}{Feature Representation Learning}}
$ X^{(t)} \leftarrow (I+L^{(t-1)})^kX^{(t-1)}$\tcp*[r]{Eq.\ref{eq:featurerefine}}
\tcp{\textcolor{violet}{Query-Support Graph Refining}}
$A^{(t)} \leftarrow \textsc{Update}(X^{(t)},B) $\tcp*[r]{Eq.\ref{eq:graphrefine}}
}
\tcp{\textcolor{violet}{Initialize Prototypes}}
$\boldsymbol{c}^{(0)}_n = \frac{1}{K}\sum_{\boldsymbol{x}'_s\in \mathcal{S}_n}\boldsymbol{x}'_s$ \tcp*[r]{Eq.\ref{eq:initproto}}
\tcp{\textcolor{violet}{Compute Propagation Matrix}}
$P = (I-\alpha L)^{-1}$\;
\For{$t=1:T_{\mathrm{PSLP}}$}{
\tcp{\textcolor{violet}{Prototype-based Soft-label Generation}}
    $Z^{(t)} \leftarrow \textsc{Generate}(X,C^{(t)},Y)$\tcp*[r]{Eq.\ref{eq:query},\ref{eq:support}}
\tcp{\textcolor{violet}{Soft-label Propagation}}
    $Z^{(t)} \leftarrow \textsc{Normalize}(PZ^{(t)})$\tcp*[r]{Eq.\ref{eq:support},\ref{eq:propagation},\ref{eq:normalize}}
    \tcp{\textcolor{violet}{Prototype Rectification}}
    $C^{(t+1)} \leftarrow (1-\beta) C^{(t)} + \beta (Z^{{(t)}})^\top X$\tcp*[r]{Eq.\ref{eq:update}}
}
 \KwRet{$\tilde{y}_q = \mathrm{arg}\max_n(Z_{qn}^{T_{\mathrm{PSLP}}})$}\tcp*[r]{\textcolor{violet}{Prediction}}

\end{algorithm}

\subsection{Comparison with Related Works}
Label propagation methods have usually been employed for transductive FSL problems. 
For example, iLPC \cite{ilpc} progressively adds the query samples with the highest confidence to the support set for subsequent iterative propagation.
AM \cite{am} introduces learnable parameters and learnable class centroids during the label propagation and optimizes them through multiple loss functions based on test time learning.
ProtoLP \cite{protolp} utilizes prototypes to construct a graph structure and optimize the projection of the label through label propagation. 
Compared with the related works above, the main differences are as follows. 
First, our proposed PSLP propagates the label of all samples to achieve robust results, which can also adapt well to the class-imbalanced setting. 
Second, PSLP utilizes soft-label propagation to gradually rectify the prototypes and assign the soft-labels to query samples, which leads to more stable and robust learning. Third, the proposed JMP captures the global relations of all samples to jointly learn better prototype and QSGraph representation. 
\externaldocument{3_method}
\section{Experiments}
\subsection{Datasets and Settings}

To verify the effectiveness of the proposed method, the experiments are conducted on four commonly used datasets: \textit{mini}-ImageNet \cite{mini}, \textit{tiered}-ImageNet \cite{tiered}, CUB-200-2011 \cite{cub}, and CIFAR-FS \cite{cifar1,cifar2}. Following the prior works \cite{hubs}, we adopt the same data partitioning, pre-processing procedures, and pre-trained backbone networks \cite{resnet,wideres} for feature extraction, i.e., ResNet-12 \cite{ICI}, ResNet-18 \cite{tim}, and WideResNet-28-10 \cite{s2m2}. The dimension of extracted features is subsequently reduced to 40 by PCA. 
We implement the experiments on 10,000 tasks with 1-shot and 5-shot settings, where the number of query samples $M$ is set to 75, and  the average classification accuracy along with 95\% confidence interval is reported for evaluation.  In the class-balanced setting, in line with \cite{ease,protolp}, the Sinkhorn-Knopp algorithm \cite{ot} is employed in Eq.\ref{eq:normalize}, and each category has $\frac{M}{N} = 15$ query samples. In the class-imbalanced setting, following \cite{atim}, query samples are generated following a Dirichlet distribution with parameter $\alpha_{dir} = 2$.
The hyperparameters are set as follows: for all tasks $B=8$, $\gamma = 10$ , $T_{\mathrm{PSLP}}=10$, for balanced setting $\alpha = 0.7$, $\beta = 0.6$, $k=4$, and for imbalanced setting $\alpha = 0.9$, $\beta = 0.2$, $k=1$. 

\begin{table*}
\centering
\caption{Comparison results of transductive FSL methods in the class-balanced setting on \textit{mini}-ImageNet, \textit{tiered}-ImageNet and CUB-200-2011 datasets. The best and second best performance are marked in bold and underline respectively. *: our implementation.}
\label{tab:main_result_balance}
\begin{adjustbox}{width=\linewidth}
    \begin{tabular}{lccllllll} 
    \toprule
         \multirow{2}{*}{Method}&  \multirow{2}{*}{Publication}&  \multirow{2}{*}{Backbone}&  \multicolumn{2}{c}{\textit{mini}-ImageNet}&  \multicolumn{2}{c}{\textit{tiered}-ImageNet}&  \multicolumn{2}{c}{CUB-200-2011}\\ 
         \cmidrule(lr){4-9}
         &  &  & 1-shot&  5-shot&  1-shot&  5-shot&  1-shot&  5-shot\\
         \midrule
EPNet \cite{epnet}& ECCV'2020& \multirow{6}{*}{ResNet-12} & 66.50\scriptsize{$\pm$0.89}&81.06\scriptsize{$\pm$0.60} &76.53\scriptsize{$\pm$0.87} & 87.32\scriptsize{$\pm$0.64}
 &82.85\scriptsize{$\pm$0.81} &91.32\scriptsize{$\pm$0.41}\\
 LR+ICI \cite{ICI}& CVPR'2020& & 66.85\scriptsize{$\pm$0.92}& 79.26\scriptsize{$\pm$0.55}& 80.79\scriptsize{$\pm$0.84}& 87.92\scriptsize{$\pm$0.50}& 88.06\scriptsize{$\pm$0.79}&92.53\scriptsize{$\pm$0.35}\\

    
iLPC \cite{ilpc}&  ICCV'2021&  &  69.79\scriptsize{$\pm$0.99}&  79.82\scriptsize{$\pm$0.55}&  83.49\scriptsize{$\pm$0.84}&  89.48\scriptsize{$\pm$0.47}&  89.00\scriptsize{$\pm$0.70}&  92.47\scriptsize{$\pm$0.35}\\
 EASE \cite{ease}& CVPR'2022
& & \underline{70.47\scriptsize{$\pm$0.30}}& 80.73\scriptsize{$\pm$0.16}& \underline{84.54\scriptsize{$\pm$0.27}}& 89.63\scriptsize{$\pm$0.15}& 80.11\scriptsize{$\pm$0.21}&93.13\scriptsize{$\pm$0.11}\\ 
ProtoLP* \cite{protolp}&  CVPR'2023&  &  70.22\scriptsize{$\pm$0.27} &  \underline{80.93\scriptsize{$\pm$0.16}} &  83.87\scriptsize{$\pm$0.25} &  \underline{89.73\scriptsize{$\pm$0.15}}
&  \underline{89.99\scriptsize{$\pm$0.20}} &  \underline{93.22\scriptsize{$\pm$0.10}}
\\ 
\rowcolor{yellow!10}\textbf{PSLP(Ours)}&- &  &  \textbf{70.94\scriptsize{$\pm$0.29}}&  \textbf{81.27\scriptsize{$\pm$0.16}}&  \textbf{84.90\scriptsize{$\pm$0.26}} &  \textbf{90.02\scriptsize{$\pm$0.15}} &  \textbf{90.38\scriptsize{$\pm$0.20}}&  \textbf{93.41\scriptsize{$\pm$0.10}} \\
\midrule
TIM \cite{tim}&  NIPS'2020
&  \multirow{7}{*}{ResNet-18}&  73.92\scriptsize{$\pm$0.27} &  85.04\scriptsize{$\pm$0.15} &  79.94\scriptsize{$\pm$0.26} &  88.53\scriptsize{$\pm$0.15} &  82.19\scriptsize{$\pm$0.26} &  90.79\scriptsize{$\pm$0.12} 
\\ 
ReRep \cite{rerep}&  ICML'2021
& & 76.54\scriptsize{$\pm$0.26} & 85.20\scriptsize{$\pm$0.14} & 82.58\scriptsize{$\pm$0.26} & 88.68\scriptsize{$\pm$0.16} & 85.36\scriptsize{$\pm$0.23} & 90.99\scriptsize{$\pm$0.12} 
\\ 
AM \cite{am}& Arxiv'2023
& & \underline{77.35} & \textbf{85.47} & \underline{83.40} & \underline{89.07} 
& \textbf{86.64}& \textbf{91.78}
\\ 
TCPR* \cite{tcpr}& NIPS'2022
& & 75.99\scriptsize{$\pm$0.26} & 84.39\scriptsize{$\pm$0.15} & 82.65\scriptsize{$\pm$0.26} & 88.26\scriptsize{$\pm$0.16} & 84.90\scriptsize{$\pm$0.23} & 90.10\scriptsize{$\pm$0.11} 
\\ 
ProtoLP* \cite{protolp}& CVPR'2023
& & 77.26\scriptsize{$\pm$0.27} & 85.07\scriptsize{$\pm$0.14} & 82.99\scriptsize{$\pm$0.26} & 88.70\scriptsize{$\pm$0.15} & 85.86\scriptsize{$\pm$0.24} & 90.52\scriptsize{$\pm$0.11}\\ 
noHub-S \cite{hubs}& CVPR'2023
& & 76.68\scriptsize{$\pm$0.28} & 84.67\scriptsize{$\pm$0.15} & 83.09\scriptsize{$\pm$0.27} & 88.43\scriptsize{$\pm$0.16} & 85.81\scriptsize{$\pm$0.24} & 90.52\scriptsize{$\pm$0.12} 
\\ 

\rowcolor{yellow!10}\textbf{PSLP(Ours)}& -& & \textbf{77.65\scriptsize{$\pm$0.28}}& \underline{85.32\scriptsize{$\pm$0.14}} & \textbf{83.65\scriptsize{$\pm$0.27}} & \textbf{88.96\scriptsize{$\pm$0.15}} 
& \underline{86.31\scriptsize{$\pm$0.24}} & \underline{91.07\scriptsize{$\pm$0.11}} 
\\ 
\midrule
 EPNet \cite{epnet}& ECCV'2020& \multirow{11}{*}{WRN-28-10} & 70.74\scriptsize{$\pm$0.85}
 &84.34\scriptsize{$\pm$0.53} &78.50\scriptsize{$\pm$0.91} & 88.36\scriptsize{$\pm$0.57}
 &87.75\scriptsize{$\pm$0.70} &94.03\scriptsize{$\pm$0.33}\\
LaplacianShot \cite{laplacian}& ICML'2020
& & 74.86\scriptsize{$\pm$0.19} & 84.13\scriptsize{$\pm$0.14} & 78.56\scriptsize{$\pm$0.21} & 87.00\scriptsize{$\pm$0.15} & 87.50\scriptsize{$\pm$0.19} & 92.30\scriptsize{$\pm$0.10} 
\\
ReRep \cite{rerep}& ICML'2021
& & 80.04\scriptsize{$\pm$0.23} & 87.64\scriptsize{$\pm$0.12} & 84.30\scriptsize{$\pm$0.25} & 90.01\scriptsize{$\pm$0.15} & 89.78\scriptsize{$\pm$0.18} & 92.20\scriptsize{$\pm$0.10} 
\\ 
iLPC \cite{ilpc}& ICCV'2021
& & 83.05\scriptsize{$\pm$0.79} & 88.82\scriptsize{$\pm$0.42} & 88.50\scriptsize{$\pm$0.75} & 92.46\scriptsize{$\pm$0.42} & 91.37\scriptsize{$\pm$0.63} & 93.93\scriptsize{$\pm$0.30} 
\\ 
EASE \cite{ease}& CVPR'2022
& & 83.00\scriptsize{$\pm$0.21} & 88.92\scriptsize{$\pm$0.13} & \underline{88.96\scriptsize{$\pm$0.26}} & \underline{92.63\scriptsize{$\pm$0.16}}
& 91.68\scriptsize{$\pm$0.19} & 94.12\scriptsize{$\pm$0.09} 
\\ 
         TCPR* \cite{tcpr}& NIPS'2022
& & 81.27\scriptsize{$\pm$0.24} & 87.80\scriptsize{$\pm$0.13} & 81.89\scriptsize{$\pm$0.27} & 87.95\scriptsize{$\pm$0.16} 
& 91.91\scriptsize{$\pm$0.18} & \underline{94.25}\scriptsize{$\pm$0.10} 
\\ 
noHub-S \cite{hubs}& CVPR'2023
& & 82.00\scriptsize{$\pm$0.26} & 88.03\scriptsize{$\pm$0.13} & 82.85\scriptsize{$\pm$0.27} & 88.31\scriptsize{$\pm$0.16} & \textbf{92.63\scriptsize{$\pm$0.18}} & \textbf{94.69\scriptsize{$\pm$0.09}}
\\ 
ProtoLP* \cite{protolp}& CVPR'2023
& & 82.53\scriptsize{$\pm$0.24} & 88.95\scriptsize{$\pm$0.13} & 88.55\scriptsize{$\pm$0.22} & 92.61\scriptsize{$\pm$0.13} 
& 91.51\scriptsize{$\pm$0.18} & 94.01\scriptsize{$\pm$0.10} 
\\ 
AM \cite{am}& Arxiv'2023
& & \textbf{83.40\scriptsize{$\pm$0.74}} & \underline{89.08\scriptsize{$\pm$0.40}} & 88.31\scriptsize{$\pm$0.73} & 92.60\scriptsize{$\pm$0.39} & 91.32\scriptsize{$\pm$0.60} & 94.14\scriptsize{$\pm$0.29} 
\\ 
PUTM \cite{putm}& ICCV'2023
& & 81.40 & 88.10 & 86.80 & 92.20 
& 91.20 & 93.70 
\\ 
\rowcolor{yellow!10}\textbf{PSLP(Ours)}& -& & \underline{83.36\scriptsize{$\pm$0.25}} & \textbf{89.27\scriptsize{$\pm$0.13}} & \textbf{89.22\scriptsize{$\pm$0.22}} & \textbf{92.82\scriptsize{$\pm$0.13}} & \underline{91.89\scriptsize{$\pm$0.18}}& 94.19\scriptsize{$\pm$0.10}\\
\bottomrule
    \end{tabular}
\end{adjustbox}

\end{table*}
\subsection{Comparison with State-of the Art}
\subsubsection{Class-Balanced transductive FSL}

The classification comparison results by different backbones on \textit{mini}-ImageNet, \textit{tiered}-ImageNet, CUB-200-2011 datasets are shown in Table \ref{tab:main_result_balance}, and the results on CIFAR-FS are shown in Table \ref{tab:cifarfs_balance}. From the tables, the following observation can be drawn: (1) The proposed PSLP outperforms the prototype-based methods, including EASE \cite{ease}, TIM \cite{tim}, PUTM \cite{putm}, which demonstrates the effective combing label propagation with prototype learning. In addition, the proposed joint message passing utilizes an efficient approach that can improve the feature representation ability. (2) Our PSLP obtains a better accuracy compared to the label propagation-based methods, including EPNet \cite{epnet}, LaplacianShot \cite{laplacian}, iLPC \cite{ilpc}, ProtoLP \cite{protolp}. This evidently demonstrates the effectiveness of soft-label propagation for updating the prototypes and labels. Compared to iLPC and ProtoLP, our approach simplifies the model iteration steps and achieves consistent improvements of approximately 0.3\% in each setting. (3) In comparison with the methods that require excessive learnable parameters to train through backpropagation, such as AM \cite{am} and noHub-S \cite{hubs}, the proposed PSLP achieves competitive results. Compared with the nohub-S and AM, our method is more efficient and achieves higher  inference performance.

\begin{table}
\caption{Comparation result of transductive FSL methods in the class-balanced setting on Cifar-FS dataset. *: our implementation.}
    \centering
    \begin{adjustbox}{width=\linewidth}
    \begin{tabular}{lcll}
    \toprule
     \multicolumn{4}{c}{Cifar-FS}\\
     \midrule
         Method&  Backbone & 1-shot&  5-shot\\
\midrule
 LR+ICI \cite{ICI}
& \multirow{5}{*}{ResNet-12}& 75.36\scriptsize{$\pm$0.97} &84.57\scriptsize{$\pm$0.57} 
\\
 iLPC \cite{ilpc}
& & 77.14\scriptsize{$\pm$0.95} &85.23\scriptsize{$\pm$0.55} 
\\
 EASE \cite{ease}
& & \underline{78.41\scriptsize{$\pm$0.29}} &\underline{85.67\scriptsize{$\pm$0.11}} 
\\
 ProtoLP* \cite{protolp}
& & 78.08\scriptsize{$\pm$0.24} &85.66\scriptsize{$\pm$0.17} 
\\
\rowcolor{yellow!10}\textbf{PSLP(Ours)}
&  &\textbf{78.90\scriptsize{$\pm$0.29}}&  \textbf{86.02\scriptsize{$\pm$0.17}}\\
\midrule
         S2M2-R \cite{s2m2}
&  \multirow{8}{*}{WRN-28-10}&  74.81\scriptsize{$\pm$0.19} &  87.47\scriptsize{$\pm$0.13} 
\\
         LR+ICI* \cite{ICI}
&  &  84.88\scriptsize{$\pm$0.79}&  89.75\scriptsize{$\pm$0.48}
\\
         iLPC \cite{ilpc}
&  &  86.51\scriptsize{$\pm$0.75} &  90.60\scriptsize{$\pm$0.48}
\\
         EASE \cite{ease}
&  &  87.60\scriptsize{$\pm$0.23} &  90.60\scriptsize{$\pm$0.16}
\\
         ProtoLP* \cite{protolp}
&  &  \underline{87.51\scriptsize{$\pm$0.23}} &  90.68\scriptsize{$\pm$0.15}
\\
         AM \cite{am}
&  &  86.91 &  \underline{90.80} 
\\
         PUTM \cite{putm}
&  &  86.40 &  90.40 
\\
\rowcolor{yellow!10}\textbf{PSLP(Ours)}
& & \textbf{87.87\scriptsize{$\pm$0.23}} &\textbf{90.88\scriptsize{$\pm$0.15}}
\\
\bottomrule
    \end{tabular}
    \end{adjustbox}
    \label{tab:cifarfs_balance}
\end{table}

\subsubsection{Class-Imbalanced transductive FSL}

\begin{table}
\caption{Comparison result of transductive FSL methods in class-imbalanced setting on mini-ImageNet and tiered-ImageNet datasets. *: our implementation. 
}
    \centering
    \begin{adjustbox}{width=\linewidth}
    \begin{tabular}{lcccc}
    \toprule
         \multirow{2}{*}{Method}&  \multicolumn{2}{c}{\textit{mini}-ImageNet}&  \multicolumn{2}{c}{\textit{tiered}-ImageNet}\\
         \cmidrule(lr){2-5}
         &  1-shot&  5-shot&  1-shot& 5-shot \\
          \midrule
         PT-MAP \cite{ptmap}
&  63.2 &  69.3 &  67.8 & 73.2 \\
         LaplacianShot \cite{laplacian}
&  69.8 &  83.6 &  72.6 & 87.7 \\
         TIM \cite{tim}
&  70.9 &  83.2 &  77.8 & 88.3 \\
         BD-CSPN \cite{bdcspn}
&  71.2 &  83.4 &  76.3 & 87.2 \\
         $\alpha$-TIM \cite{atim}
&  70.9 &  84.9 &  77.7 & 89.9 \\
         iLPC \cite{ilpc}
&  \underline{74.3} &  83.3 & \underline{81.6}& 89.0 \\
 BAVAR \cite{bavar}
& 74.2 & 85.6 & 80.6 &89.3 \\
 AM \cite{am}
& 72.0 & 85.6 & 78.9 &88.7 \\
 ProtoLP* \cite{protolp}
& 73.7 & 85.4 & 81.0 &89.0 \\
         PUTM \cite{putm}
&  73.8 &  \underline{85.7} &  81.1 & \underline{90.4} \\
\rowcolor{yellow!10}\textbf{PSLP(Ours)}
& \textbf{75.9} &  \textbf{89.2} &  \textbf{82.7} & \textbf{92.5} \\
\bottomrule
    \end{tabular}
    \end{adjustbox}
    
    \label{tab:mini_imbalance}
\end{table}

\begin{table}
 \caption{Comparison result of transductive FSL methods in the class-imbalanced setting on CUB-200-2011 dataset. *: our implementation. 
 }
    \centering
    \begin{tabular}{lcccc}
    \toprule
    \multicolumn{5}{c}{CUB-200-2011}\\
    \midrule
          \multirow{2}{*}{Method}&  \multicolumn{2}{c}{ResNet-18}&  \multicolumn{2}{c}{WRN-28-10}\\
          \cmidrule(lr){2-5}
 & 1-shot & 5-shot & 1-shot & 5-shot \\
 \midrule
          PT-MAP \cite{ptmap}
&  63.2&  69.3&  67.8 & 73.2 
\\
          LaplacianShot \cite{laplacian}
&  73.7&  87.7&  84.2 & 91.4 
\\
          TIM \cite{tim}
&  74.8&  86.9&  82.8 & 90.5 
\\
          BD-CSPN \cite{bdcspn}
&  74.5&  87.1&  85.1 & 91.5 
\\
          $\alpha$-TIM \cite{atim}
&  75.7&  89.8&  83.6 & \underline{92.8}
\\
          iLPC \cite{ilpc}
&  78.7&  86.9&  \underline{86.4} & 90.8 
\\
          BAVAR \cite{bavar}
&  \textbf{82.0}&  \textbf{90.7}&  85.7 & 90.8 
\\
  AM \cite{am}
& 78.6& 89.9& -&-
\\
  ProtoLP* \cite{protolp}
& 76.8& 86.4& 82.3 &90.6 
\\
          PUTM \cite{putm}
& \underline{78.9}&  89.3&  \textbf{86.9} & 92.2
\\
\rowcolor{yellow!10}\textbf{PSLP(Ours)}
&  \underline{78.9}&  \underline{90.2}&  86.2& \textbf{93.3}
\\
\bottomrule
    \end{tabular}
   
    \label{tab:cub_imbalance}
\end{table}
In Table \ref{tab:mini_imbalance} and Table \ref{tab:cub_imbalance}, we present the accuracy comparison between our PSLP and SOTA methods in the class-imbalanced setting. From the results, we can note that (1) Compared to the general transductive FSL methods, such as TIM \cite{tim}, ProtoLP \cite{protolp}, and BD-CSPN \cite{bdcspn}, our method achieves state-of-the-art performance, demonstrating its effectiveness and robustness under class-imbalanced setting. (2) In comparison with the work specifically addressing the class imbalance in transductive FSL, such as $\alpha$-TIM \cite{atim}, PUTM \cite{putm}, BAVAR \cite{bavar}, and AM \cite{am}, our method provides higher accuracy on \textit{mini}-ImageNet and \textit{tiered}-ImageNet, further verifying the robustness of proposed PSLP. This is because PSLP can adaptively adjust the strength of label propagation for each class via soft labels of query samples. (3) Compared to the methods that rely on prior distributions of samples for each class, our method yields competitive results. For example, 
Compared to iLPC \cite{ilpc}, which introduces label propagation with a class-balanced prior, our PSLP achieves up to 5.9\% performance improvement. Compared with BAVER, which also relies on a class-imbalance prior, our method obtains superior results on \textit{mini}-ImageNet and \textit{tiered}-ImageNet and competitive results on CUB dataset, further proving the advantage of freedom from prior assumptions of our method.


\subsection{Ablation Experiments}

\subsubsection{Components Analysis}
To verify the effectiveness of our proposed JMP and PSLP, we conduct the ablation studies. The experimental results are shown in Table \ref{tab:ablation}. We can observe that (1) The proposed JMP achieves better results on different datasets with different settings as it jointly learns improved prototype representations and a more robust query-support graph structure.
(2) Adopting the proposed PSLP instead of the vanilla label-propagation yields a significant improvement in accuracy. This is attributed to PSLP, which alternately optimizes the prototypes and labels through soft-label propagation and provides a more accurate estimation of the prototypes and labels for each sample. (3) When the JMP and PSLP are involved, further accuracy improvement is achieved.


\begin{table}
    \caption{Component ablations of the proposed method on WRN-28-10 in balanced setting. JMP and PSLP represent joint message passing and prototype-based soft-label propagation respectively. }
    \centering
    \begin{tabular}{cccccc}
    \toprule
         \multicolumn{2}{c}{Component}&  \multicolumn{2}{c}{\textit{mini}-ImageNet}&\multicolumn{2}{c}{\textit{tiered}-ImageNet}\\
         \cmidrule(lr){1-2}\cmidrule(lr){3-6}
 JMP& PSLP& 1-shot&5-shot&1-shot&5-shot\\
 \midrule
  & &                     75.29& 87.65& 82.12 & 91.44\\
 \checkmark& &            76.70& 89.02& 83.46 & 92.62\\
 &  \checkmark&           83.24& 89.25& 89.19 & 92.79\\
 \checkmark&  \checkmark& 83.36& 89.27& 89.22 & 92.82\\
 \bottomrule
    \end{tabular}

    \label{tab:ablation}
\end{table}

\subsubsection{Sensitivity to Hyperparameters}

\begin{figure}
    \centering
    \includegraphics[width=\linewidth]{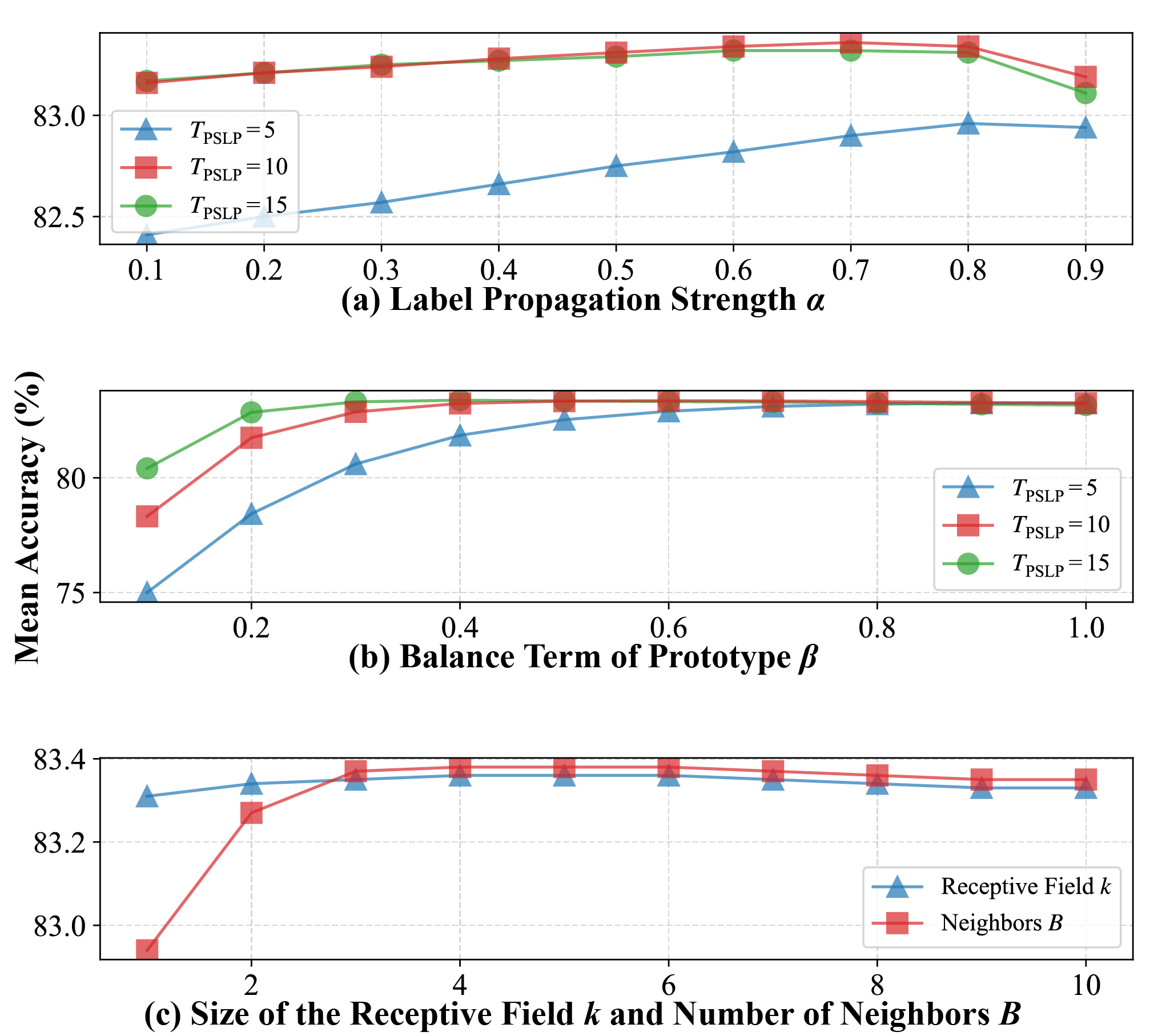}
    \caption{Mean Accuracy (\%) of 5-way 1-shot tasks with different hyperparameters on \textit{mini}-ImageNet. 
    }
    \label{fig:hyperparameters}
\end{figure}
To further figure out the influence of hyperparameters, we experiment with different values of hyperparameters on \textit{mini}-ImageNet dataset. The hyperparameters are $k$, $B$, $\alpha$, and $\beta$, which represent the size of the receptive field, the number of neighbors for each sample in JMP, the strength of label propagation, and the balance term of prototypes in PSLP, respectively. The results are shown in Figure \ref{fig:hyperparameters}. From the figure, we can observe that it leads to a decrease in accuracy when $\alpha$ and $\beta$ are too low, which can be alleviated by increasing the number of iterations. Furthermore, although $k$ and $B$ are low, the sufficiently aggregated information from similar samples can be obtained by JMP, proving the feasibility of sparse graph construction. 
Overall, our method is less affected by hyperparameters choices, with accuracy fluctuating within 0.5\% for most choices, which demonstrates that our method is not sensitive to the choice of these hyperparameters.

\subsubsection{Robustness in Real-world Scenarios}

\begin{table}
\caption{Mean accuracy (\%) of 5-way 5-shot tasks with different query numbers $M$ in the class-imbalanced setting on \textit{mini}-ImageNet dataset. * : our implemenetation.}
    \centering
    \begin{adjustbox}{width=\linewidth}
    \begin{tabular}{lccccc}
    \toprule
         \multirow{2}{*}{Method}&  \multicolumn{5}{c}{Query Number}\\
         \cmidrule(lr){2-6}
         &  15&  45&  75&  105& 135\\
         \midrule
         $\alpha$-TIM* \cite{atim}&  83.68& 85.02&  85.67&  \underline{86.10}& \underline{86.18} \\
         ProtoLP* \cite{protolp}&  \underline{97.15} &  \underline{88.89} &  85.37&   83.35&82.24 \\
         PUTM* \cite{putm}&  84.77 &  85.04 &  85.14 &  85.03 &85.04 \\
         \textbf{PSLP(Ours)} &  \textbf{98.32}& \textbf{ 91.53} &  \textbf{89.23} &  \textbf{87.77} & \textbf{86.96 }\\
         \bottomrule
    \end{tabular}
 \end{adjustbox}
    
    \label{tab:robustness}
\end{table}
A crucial aspect for practical applications is the robustness of method under different class-imbalanced setting. To evaluate the robustness of the proposed PSLP, we vary the number of query samples for each class or the class distribution as it is not always known for real-world scenarios. From Table \ref{tab:robustness}, as $M$ decreases, label-class samples dominate, allowing methods based on label propagation, such as ProtoLP and our PSLP, to achieve higher accuracy. 
Compared to ProtoLP \cite{protolp}, our method significantly improves accuracy. The $\alpha$-TIM \cite{atim} and PUTM \cite{putm} give lower accuracy, which is mainly due to the insufficient number of samples available for test-time training. Although the accuracy of our method declines slightly when the proportion of labeled samples decreases as $M$ increases, it is still higher than the compared methods. 
This verifies the robustness of our method in real-world scenarios.

\subsubsection{Inference Speed}
\label{sec:speed}
\begin{table}
 \caption{Inference time (s) for 1000 tasks of 5-way 5-shot on \textit{mini}-ImageNet by CPU. Tasks are build with the features extracted by backbones. }
    \centering
    \begin{tabular}{ccccc}
    \toprule
         Backbone&  $\alpha$-TIM&  PUTM&  ProtoLP& Ours\\
         \midrule
ResNet12&90.58 &454.93 &7.16 &1.54\\
WRN-28-10&  119.27 &  477.29&  9.02& 1.86\\
         \bottomrule
    \end{tabular}
   
    \label{tab:inferencespeed}
\end{table}
To validate the efficiency of our method, we implement the $\alpha$-TIM \cite{atim},  ProtoLP \cite{protolp}, PUTM \cite{putm} on the same CPU (Intel I7-13700F). Table \ref{tab:inferencespeed} demonstrates the inference time cost for 1000 few-shot tasks.
From the table, our method exhibits an evident advantage in inference time in comparison with the compared methods. It is believed because: (1) Both $\alpha$-TIM and PUTM need to optimize the learnable parameters during inference, (2) ProtoLP requires the construction of  graph iteratively and has more complex iteration procedure. In contrast, the proposed PSLP involves no learnable parameters. The inference speed is fast and converges after only a few iterations.

\section{Conclusion}

In this paper, we integrate the advantages of prototype-based and label propagation-based methods and develop an efficient and robust transductive FSL method, termed Prototype-based Soft-label Propagation (PSLP). Our proposed PSLP propagates the labels of all samples in which the soft labels of query samples are assigned by prototypes. By alternatively updating the prototypes with propagated labels and assigning new soft labels to query samples for propagation, the prototypes and labels are progressively estimated. The soft-label propagation adaptively adjusts the propagation strength of each class, which is robust to address the class-imbalanced setting problem. Moreover, we further propose the joint message passing to utilize the global information of all samples to refine feature representation and query-support graph structure.
Notably, the proposed method is parameter-free and efficient, facilitating the end-to-end inference on the CPU. Our method achieves competitive results compared to the state-of-the-art methods under various settings. 
{
    \small
    \bibliographystyle{ieeenat_fullname}
    \bibliography{main}
}


\end{document}